\documentclass[a4paper,12pt]{article}
\usepackage[english]{babel}
\usepackage{graphicx}
\usepackage{float}
\usepackage{amsfonts}
\usepackage{amsmath}
\usepackage{authblk}
\usepackage{booktabs} 

\DeclareMathOperator*{\E}{\mathbb{E}}

\begin{document}

\title{Domain Adaptation in Highly Imbalanced and Overlapping Datasets}
\author{Ran Ilan Ber}
\author{Tom Haramaty}
\affil{K Health, New York, NY}

\maketitle

\begin{abstract}
	In many machine learning domains, datasets are characterized by highly imbalanced and overlapping classes. Particularly in the medical domain, a specific list of symptoms can be labeled as one of various different conditions. Some of these conditions may be more prevalent than others by several orders of magnitude.
	Here we present a novel unsupervised domain adaptation scheme for such datasets. The scheme, based on a specific type of Quantification, is designed to work under both label and conditional shifts.
	It is demonstrated on datasets generated from electronic health records and provides high quality results for both Quantification and Domain Adaptation in very challenging scenarios.\\
	Potential benefits of using this scheme in the current COVID-19 outbreak, for estimation of prevalence and probability of infection are discussed.
\end{abstract}

\section{Background}
Our work relates to several research topics including Quantification and Domain Adaptation (DA).

\paragraph{Quantification.}
Quantification \cite{gonzalez2017review} is the task of estimating label distribution in a target dataset using a classifier that was trained on a source dataset. It is usually assumed that the conditional probabilities of input given a label remain fixed.
Quantification can be used to answer very interesting questions. For example, estimating the number of people infected with a disease in a given population.
The most straightforward Quantification method is based on a confusion matrix which relates the label distribution estimated by the classifier and the actual label distribution \cite{mclachlan1988mixture,mclachlan1992discriminant, lipton2018detecting}. In this paper we follow this approach.

\begin{figure}
	\centering
	\includegraphics[width=\linewidth]{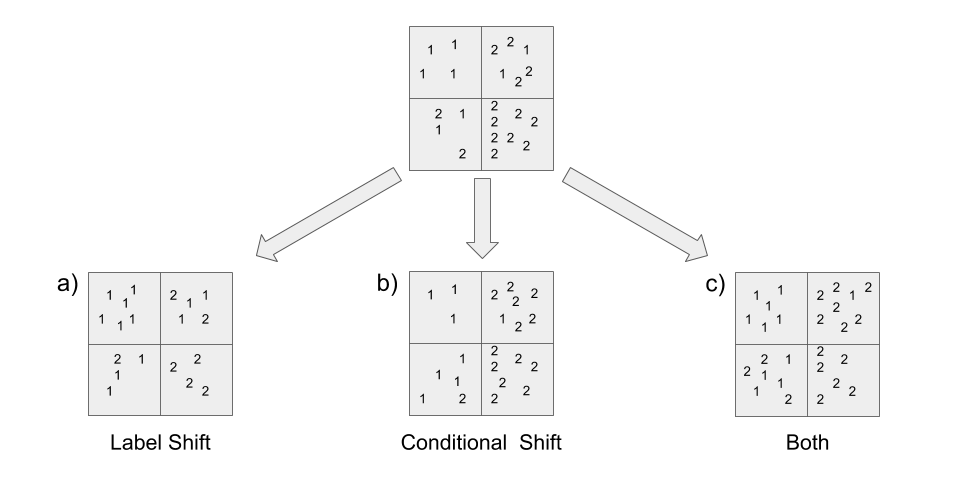}
	\caption{three types of DA: datasets with two classes denoted as '1' and '2' are illustrated. The data points are embedded in a space, separated into four subspaces, roughly corresponding to the classifier's ``resolution'' in the input space. On top there is a source data set and below are possible target datasets: a) label shift: $p(y)$ differs but $p(\mathbf{X}|y)$ remains fixed. b) conditional shift: $p(y)$ remains fixed but $p(\mathbf{X}|y)$ differs. c) label and conditional shifts: $p(y)$ and $p(\mathbf{X}|y)$ differ.}
	\label{Fig_types_of_da}
\end{figure}

\paragraph{Domain Adaptation.}
DA \cite{redko2019advances} is the task of adjusting a model from a source dataset to a different, yet related, target dataset. Unlike Quantification, DA focuses on optimizing classification.\\
DA methods differ in supervision resources. Some methods are supervised, some are semi-supervised and others are unsupervised \cite{Kouw2019generalization}. This work will focus on the unsupervised case.\\
We wish to characterize 3 types of DA (see Figure~(\ref{Fig_types_of_da})).\\
1) label distribution, $p(y)$, differs between source and target, but conditional probabilities of input given a label, $p(\mathbf{X}|y)$, remain fixed. This scenario is known in the literature as a \textit{label shift} (Figure~(\ref{Fig_types_of_da}a)).\\
2) label distribution remains fixed, but conditional probabilities of input given a label, differ. This scenario is known in the literature as a \textit{conditional shift} (Figure~(\ref{Fig_types_of_da}b)).\\
3) both label distribution and conditional probabilities of input given a label, differ (Figure~(\ref{Fig_types_of_da}c)). This is the most challenging type of DA. A notable example of a method tackling this task using adversarial techniques in the computer vision domain is presented in \cite{chidlovskii2019using}. Here we address this task from a different approach, which we believe is more suitable for tabular medical datasets.\\

\section{Introduction}
In datasets characterized by overlapping classes (the same input may be labeled in different probabilities to different classes), there are significant ``confusion effects'' in which an increase in a certain label's probability results in the model estimating an increase in other labels' probabilities, even when the best possible models are used.
These effects become even more dominant when classes are imbalanced since, given an overlap between a rare class and a prevalent class, models will learn to assign most of the probability from data points associated with the rare class to the prevalent class.
As most DA methods were developed under the assumption that classes are separable, they don't take these effects into account, which may lead to poor results in such datasets.\\
However, DA is in fact very much needed in these datasets, even when only minor shifts are involved.
Due to the overlapping, the classifier will generally not learn definite $p(y|\mathbf{X})$, and so it will heavily rely on $p(y)$ that it has learned from the source dataset. However, due to the imbalance, even a small label shift will cause significant differences in $p(y)$, especially for small classes, thus leading to poor results.

A good example for such datasets can be found in the medical field. Consider the task of classifying a list of symptoms to a condition \cite{wu2010prediction,shickel2017deep}.
In this scenario, as symptoms alone cannot usually be used to pinpoint a single condition, the best a symptom-based classifier can do is assign reasonable probabilities to a number of conditions related to the reported symptoms. Some of these conditions may be more prevalent than others by several orders of magnitude. In fact, some conditions may be so rare and overshadowed by other conditions, that none of the cases will be classified to them (it is useful to include such conditions in datasets in order to generate a differential diagnosis from the classifier's outcome, for example).
A label shift is expected in this scenario as condition prevalence often varies significantly between populations and time periods.
Less intuitively, a conditional shift is also expected. Symptom distribution given conditions, tend to change between populations and time periods due to differences in reporting methods (e.g., two different machine-patient dialogue algorithms that are used to extract symptoms).\\

Here we present a novel scheme for unsupervised DA which is tailored to imbalanced and overlapping datasets and works under label and conditional shifts.
Before we do so, we introduce a Quantification method which is known in the literature but is under-explored. We re-derive it and analyze it in light of a more common approach, in order to set the building blocks for the remainder of the work.
Then, we shortly describe a standard technique which employs Quantification to perform DA under a label shift.
Finally, we utilize the Quantification and DA methods mentioned above to derive our novel DA scheme. The scheme is discussed theoretically and demonstrated on datasets, generated from electronic health records (EHRs), involving the classification of a list of symptoms to a condition.

\paragraph{Outline}
The remainder of this article is organized as follows:
in Section~\ref{DA} we present our DA scheme and explain it theoretically,
Section~\ref{Experiments} elaborates on experiments performed with this scheme and Section~\ref{discussion} discusses our findings.

\section{The Domain Adaptation Scheme}\label{DA}
\subsection{Quantification under a label shift}\label{Quantification Using the Soft CM}
Consider a dataset in which classes are highly imbalanced and overlapping, and a classifier trained on this dataset.
We would like to use this classifier in order to estimate the label distribution in an unlabeled target set (at this point it is assumed there is no conditional shift). When using the classifier as is, for reasons explained above, the estimation is expected to be very poor.

Denoting the label distribution as $P_y$, and the probability that the classifier assigned to class $y$ given input $\mathbf{X}$ as $\hat{y}$, the classifier estimation of the label distribution is given by $\hat{P}_y=\E [\hat{y}]$.\\
In order to quantify the classifier's confusion we define the \textit{soft} confusion matrix
\begin{equation}
C_{yy'}=\E[\hat{y}|y'], \label{soft confusion matrix}
\end{equation}
Next, plugging the definitions above we obtain $C P = \hat{P}$, and so
\begin{equation}
P = C^{-1} \hat{P} \label{prev estimated prev relation}.
\end{equation}
Note that $C$ can be estimated from the source dataset, while $\hat{P}$ can be estimated from the target dataset using a classifier trained on the source dataset. Hence, Eq. (\ref{prev estimated prev relation}) provides a recipe for performing Quantification, i.e. for evaluating label distribution in a target dataset, without using labels from the target dataset.
The classifier does not have to be calibrated. In fact, this procedure may be used to calibrate classifiers on the source dataset as well.
Note that this procedure works only if $C$ is nonsingular. $C$ will be singular if $\hat{P}_y = 0$ for one or more classes, or if the expected class distribution given a certain label is a linear combination of the expected class distributions given other labels. $C$ is very unlikely to be singular, but the closer it gets to singularity the more sensitive to noise it becomes.

\begin{figure}
	\centering
	\includegraphics[width=\linewidth]{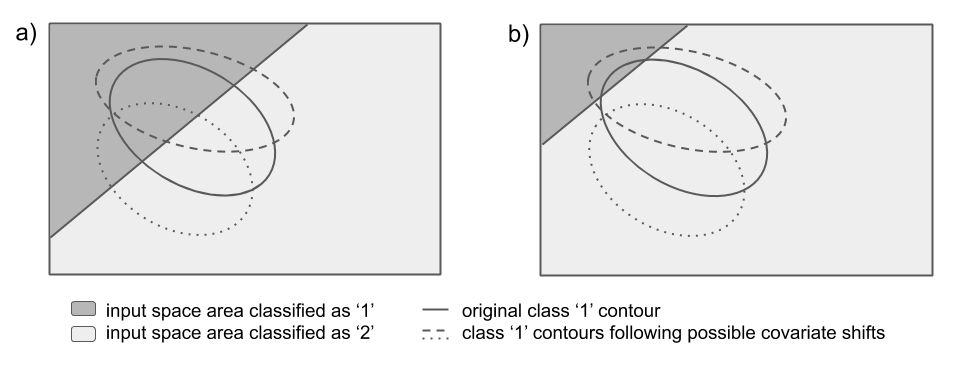}
	\caption{sensitivity of Quantification to conditional shifts. Consider a rare class '1' and a prevalent class '2'. a) overlap between classes is mild. The amount of data points from class '1' classified correctly in the source dataset is in the same order of magnitude as in the two conditional-shifted targets (marked with dashed and dotted lines). b) overlap between classes is extreme. In the target dataset marked with the dashed line, the number of data points from class '1' classified correctly is much larger than that of the source (leading to poor results), while in the target dataset marked with the dotted line, non of the data points are classified as '1' (leading to an inability to perform the process).}
	\label{Fig_sensitivite_to_cond_shift}
\end{figure}

The procedure described above is known in the literature but is usually performed with a hard confusion matrix rather than a soft one (in the notation we use it means replacing $\hat{y}$ with its binary equivalent: $1$ for the $\hat{y}$ with the highest probability and $0$ elsewhere).\\
The soft confusion matrix is mostly mentioned in the literature as a replacement for the hard confusion matrix in case it is singular \cite{lipton2018detecting,garg2020unified}.
Here, we claim that in highly overlapping and imbalanced datasets this should be the confusion matrix used for several reasons:\\
1) \textit{singularity}: if non of the data points in the source dataset are classified as a certain class, the hard confusion matrix becomes singular. As mentioned above, this often happens in such datasets since rare classes tend to be underestimated \cite{forman2008quantifying}.\\
2) \textit{noise}: comparing the expected noise of Quantification in the limit of extremely imbalanced and overlapping classes, we find that the noise in the soft approach is smaller. Consider, for simplicity, datasets with two classes, one rare and one prevalent. For the rare class, whose probability is $p$, only $\epsilon \ll 1$ of its instances are classified correctly ($\epsilon$ is determined by $p$ and the nature of the overlap). It is easy to show that under reasonable assumptions, the relative noise of the hard approach for the rare class probability estimation will scale like $\mathcal{O}(1/\sqrt{\epsilon p})$, while that of the soft estimation will scale like $\mathcal{O}(1/\sqrt{p})$.\\
3) \textit{sensitivity to conditional shift}: using the hard approach, when classes are not extremely imbalanced and overlapping, minor conditional shifts will have a small effect on Quantification results (Figure~(\ref{Fig_sensitivite_to_cond_shift}a)). However, the more imbalance and overlap exists, the more Quantification is sensitive to minor conditional shifts (Figure~(\ref{Fig_sensitivite_to_cond_shift}b)). This is due to the fact that rare classes are underestimated. Once $\epsilon \ll 1$ of the data points associated with a rare class are classified correctly, even a very small conditional shift (compared to the scale of the class contour) can lead, on one hand to an $\epsilon' \gg \epsilon$, and on the other hand to $\epsilon'=0$ (which results in a singular confusion matrix). The changes in the soft approach will be much smoother.\\
4) \textit{generalization to DA with conditional shifts}. An explanation will follow in Section~\ref{Generalizing to CS}.

\subsection{From Quantification to DA}
So far, we have shown how to estimate the label distribution in a target dataset.
In order to calibrate the classifier to the target set, and thus improve its classification quality, a known procedure is followed \cite{saerens2002adjusting}. Recall that every (calibrated) classifier can be described by Bayes Law,
\begin{equation}
\hat{y}^{s} \propto p(\mathbf{X}^s | y^{s}) P^{s}_y,
\end{equation} 
where the superscript 's' emphasizes that these quantities are calibrated to the source dataset. Once we have an estimation of class distribution in the target dataset, it may be used to calibrate the classifier as follows:
\begin{equation}
\hat{y}^{t} \approx \hat{y}^{s} \cdot \frac{P^{t}_y}{P^{s}_y},
\end{equation}
(the subscript 't' stands for 'target') where, again, we have assumed no conditional shift.\\
This implies that every classifier trained on the source dataset can be calibrated to a target dataset, in which class distribution is potentially very different than that of the source, with only a small number of numerical operations.

\subsection{Adding a conditional shift}\label{Generalizing to CS}
No conditional shift has been assumed above.
However, in many cases this assumption is not realistic, hence we wish to relax it.
In order to do so, we divide the input space into subspaces and apply the Quantification and calibration schemes for each subspace separately.
This will work if for every subspace $\sigma$ and every class $y$ the following condition is satisfied:
\begin{equation}
\frac{p(\mathbf{X}^t_\sigma|y^t)}{p(\mathbf{X}^s_\sigma|y^s)} = \lambda_{y\sigma}, \quad \forall \mathbf{X}_\sigma \in \sigma \label{condition for DA to succeed}
\end{equation}
where $\lambda_{y\sigma}$ are constants. This is because when this condition is satisfied, the conditional shift translates in every single subspace into a label shift,
\begin{equation}
\hat{y}^t_\sigma \propto p(\mathbf{X}^t_\sigma|y^t)P^t_y = p(\mathbf{X}^s_\sigma|y^s) P^t_{y\sigma},
\end{equation}
where $P^t_{y\sigma} = \lambda_{y\sigma} P^t_y$ is the probability of class $y$ in $\sigma$ in the target dataset.
In order to satisfy this condition, the shifts in $p(\mathbf{X}|y)$ need to be on a larger scale than the size of the subspaces. We are aware of two plausible scenarios where this happens:\\
1) \textit{sub-classes}: consider a scenario in which classes are composed of separated sub-classes, and where the conditional shift results only from different weighting of the sub-classes. In this scenario, if the space division corresponds to the sub-classes, condition~\ref{condition for DA to succeed} will be satisfied by definition.\\
In the medical domain, such a scenario is plausible as conditions are often composed of ``sub-conditions''. For example, viral upper respiratory infections and bacterial upper respiratory infections are both composed of several ``sub-conditions'', such as conditions localized to the nasal passages, conditions of pharyngeal origin and others.
For different populations or different points in time, the ratio between various ``sub-conditions'' may change (in a different manner for each condition), leading to a conditional shift of the kind this approach can handle.\\
2) \textit{conditional independence of features with respect to the classes}: denoting the input vector as $\mathbf{X}=(X_1,X_2,...,X_F)$ (where $F$ is the number of features), assuming it can be split into $V>1$ vectors $\mathbf{X}^v$ which satisfy conditional independence with respect to the classes, we get $p(\mathbf{X}|y)=\prod_{v \in V} p(\mathbf{X}^v|y)$.
Without loss of generality, suppose the conditional shift results from changes in $p(\mathbf{X}^1|y)$, then, assuming categorical features, by dividing the input space in a way which corresponds to the different possible combinations of $\mathbf{X}^1$, conditional independence assures that condition~\ref{condition for DA to succeed} is satisfied, as
\begin{equation}
\hat{y}^t_\sigma \propto \prod_v [p(\mathbf{X}^{v,t}_\sigma|y^t)]\cdot P^t_y = \prod_{v \neq 1} [p(\mathbf{X}^{v,t}_\sigma|y^t)]\cdot P^t_y = p(\mathbf{X}^{s}_\sigma|y^s)P^t_{y\sigma}.
\end{equation}
In the medical domain, conditional independence of features (or groups of features) is often satisfied or at least approximated. If needed, the groups $v$ may be evaluated from the source dataset.\\
Suppose a classifier is trained with data obtained using a certain dialogue algorithm that asks users questions regarding (categorical, usually even binary) symptoms. If at some point in future this algorithm is changed, this will most definitely lead to shifts in $p(X_i|y)$ for some features $X_i$.
It is impossible to know in advance which features were changed due to the unknown label shift, however, the dialogue algorithm and global feature distribution may provide leads.

We note that in order to work separately on different subspaces, soft confusion matrices must be used, as all problems mentioned in Section ~\ref{Quantification Using the Soft CM} regarding hard confusion matrices worsen when we apply the Quantification and calibration schemes on small subspaces. This is because in certain subspaces, some rare classes will be even rarer.
In this context, it is worth mentioning that even though we assume an extremely overlapping dataset, some classes may have no support in some subspaces (meaning that in the source dataset, none of the data points in the subspace are labeled to these classes). In order to avoid singularity (or poor results) one should include in each subspace only classes that are supported in it.

Importantly, while bias decreases with the number of subspaces, variance will increase. This occurs due to the following:\\
1) the fewer data points that exist in a subspace, the more the scheme is vulnerable to noise. Denoting the number of subspaces as $N_\sigma$, and assuming they all contain approximately the same number of data points, the noise scales like $\mathcal{O}(\sqrt{N_\sigma})$.\\
2) the smaller the subspace, the more uniform the predictions in the subspace become, hence, the closer the confusion matrix gets to singularity. In the limit where the subspace area goes to zero, all predictions in the subspace will be identical and the confusion matrix will become singular.\\
Therefore, the space cannot be divided into arbitrarily small subspaces.

\section{Experiments}\label{Experiments}
We conducted two experiments with our scheme. Both involved the same data source and classifier:
\paragraph{Data source}
The data used is based on EHR data from Maccabi HMO (Israel) encompassing visits of adults aged 18 years and older to their family physician.
Raw EHR cases were transformed into a vector of 1221 binary symptoms (reported vs not reported) which were then fed to the classifier along with sex and age (age was the only non-binary feature). The classifier classified each of them as one of 82 possible conditions.
Data was divided randomly to source and target datasets.
Following source-target splitting, the source dataset was altered by excluding cases. It was decided to alter the source (rather than the target) as it simulates a more realistic scenario. Details pertaining to the alterations are discussed below, separately for each experiment.
\paragraph{Classifier architecture}
The classifier was implemented by a 4-layer neural net. ReLU activation was used twice for non-linearity. Batch Normalization was used twice as well. The last layer was followed by a Softmax activation to ensure that output values are all positive and sum to one.
In training, Dropout was applied to avoid overfitting and cross-entropy loss was used (along with the Softmax activation) to assign the classifier a probabilistic nature.
\paragraph{}
Described below are the experiments we performed.

\subsection{Quantification under a label shift}
To this end, each condition was separately chosen to keep a certain ratio of its original data points. The ratio was selected from a uniform distribution within the range $[0.2-1.0]$.
Excluded data points were chosen randomly.
The process culminated in approximately 11.3M data points in the source dataset and 1.9M in the target.

\begin{figure}
	\centering
	\includegraphics[width=\linewidth]{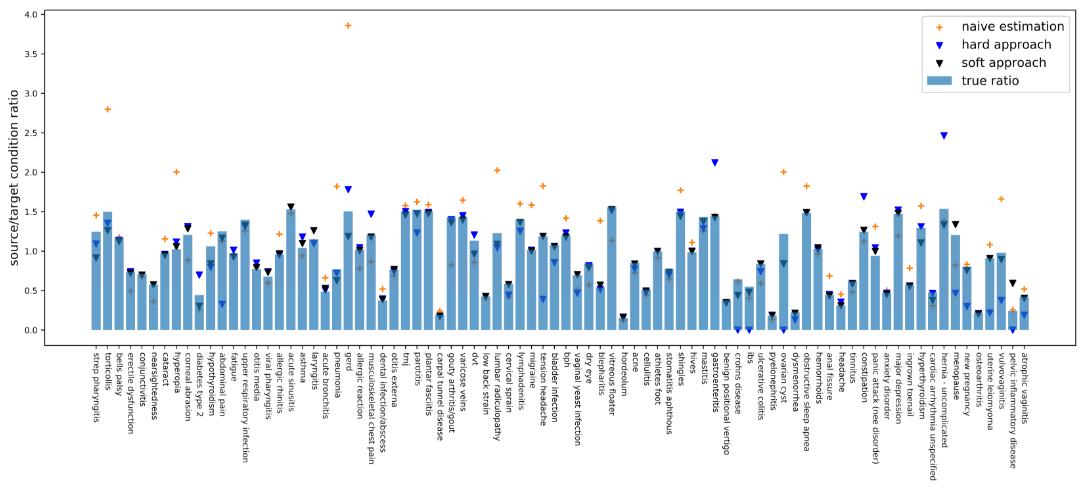}
	\caption{using different approaches to estimate the ratio between condition distribution in the source and target datasets, under a label shift. The actual ratio is represented by the bars. For each approach we show the estimated ratio and the Quantification score, defined to be the euclidean distance (divided by the number of conditions) with respect to the actual ratio (lower is better). Orange crosses: the classifier calibrated to the source dataset (score: $0.047$), blue triangles: the hard approach (score: $0.034$), black triangles: the soft approach (score: $0.012$).}
	\label{Fig_ls_results}
\end{figure}

We compared the classifier's results without calibration to results obtained with calibrations using the hard and soft approaches.
Results are presented in Figure~(\ref{Fig_ls_results}).
It is shown that the soft approach performed considerably better than the hard approach.

\subsection{Quantification and DA under both label and conditional shifts}
To this end we have only considered 8 respiratory conditions: Strep (Streptococcal) Pharyngitis, Upper Respiratory Infection, Viral Pharyngitis, Allergic Rhinitis, Acute Sinusitis, Asthma, Influenza \& Pneumonia.
As previously ascertained, samples were excluded from the source dataset. Each condition kept a certain ratio of its original data points, such that the source dataset became relatively balanced (a ratio of approximately $1:2$ between the prevalence of the largest and the smallest classes). 
The exclusion of data points was conducted by assigning each data point with a weight determined by the values of 3 prominent symptoms: runny nose, sore throat \& cough (assigned weights were different for each condition).
We then randomly selected data points using the weights as exclusion probabilities.
This changed the conditional probability of other symptoms in the source dataset as well.
Eventually, there were approximately 798K data points in the source dataset and 287K in the target.

Results of the classifier without calibration were compared to results obtained with calibrations. First, a calibration using the hard approach, without input space division (the hard approach with input space division failed due to singular confusion matrices). Then calibrations using the soft approach, with and without input space division.
The input space division involved PCA to reduce the number of dimensions in the input space to 6, followed by K-Means to divide the space into 5 subspaces.

\begin{figure}
	\centering
	\includegraphics[width=\linewidth]{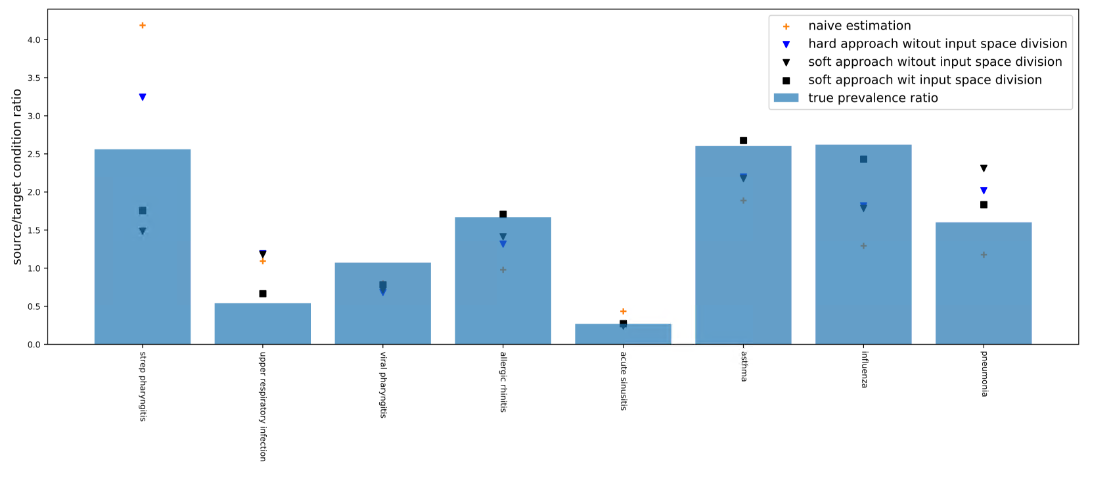}
	\caption{using different approaches to estimate the ratio between condition distribution in the source and target datasets, under both label and conditional shifts. The actual ratio is represented by the bars. For each approach we show the estimated ratio and the Quantification score, defined to be the euclidean distance (divided by the number of conditions) with respect to the actual ratio (lower is better). Orange crosses: the classifier calibrated to the source dataset (score: $0.31$), blue triangles: the hard approach without input space division (score: $0.18$),  black triangles: the soft approach without input space division (score: $0.22$), black squares: the soft approach with input space division (score: $0.12$).}
	\label{Fig_lcs_results}
\end{figure}

Quantification results are presented in Figure~(\ref{Fig_lcs_results}) and DA results are presented in Table~(\ref{Ta_lcs_results}).
Due to the conditional shift, both soft and hard approaches fail to achieve good results without dividing the input space into subspaces:
in the Quantification task, the results without dividing the input space were at least considerably better than the results obtained by a classifier calibrated to the source dataset (note that the hard approach reached better results than the soft approach, we believe this is due to the fact that the imbalance in the source dataset was not extreme).
In the DA task, the results without dividing the input space weren't significantly better than those of the classifier calibrated to the source dataset.
The soft approach with input space division was superior to the other methods and obtained very good results for both Quantification and DA.

\begin{table}
	\centering
	\begin{tabular}{lll} 
		\toprule
		
		Method & Top1 & Top3 \\ [0.5ex] 
		\midrule
		No DA                                      & 0.51          & 0.79           \\ 
		DA without input space division - hard     & 0.51          & 0.80           \\
		DA without input space division - soft     & 0.51          & 0.80           \\
		DA with input space division - soft        & \textbf{0.57} & \textbf{0.83}  \\
		\midrule
		Source dataset                             & 0.60          & 0.87           \\
		\bottomrule
	\end{tabular}\\
	\caption{classification results in a scenario involving both label and conditional shifts. Since conditional symptom distributions given a condition were altered in the source dataset, the DA methods not including the input space division failed to significantly improve results in the target dataset. On the other hand, the method including the division managed to bridge approximately half of the gap between the source and target performances in the Top3 metrics and two thirds in the Top1.}
	\label{Ta_lcs_results}
	
\end{table}

\section{Discussion}\label{discussion}
In this work we introduced a novel DA approach tailored to a challenging scenario involving both label and conditional shifts in highly imbalanced and overlapping datasets.\\
This approach is based on embedding the input into a low dimensional space, dividing it into subspaces and calibrating the classifier separately for each subspace.
We have experimented with this scheme and found that it reaches very good results for both Quantification and DA.

A simple method of dividing the input space to subspaces was examined. It involved PCA for dimensionality reduction followed by K-Means for the division. It would be interesting to investigate, both theoretically and experimentally, how different methods of space division may improve our results.
Additional improvements to our results may be achieved by using ensembles of classifiers, as suggested by \cite{lipton2018detecting}.

Finally, it would be interesting to further study this approach in light of an Active Learning scheme. The relation between Active Learning and DA has been studied extensively \cite{csurka2017comprehensive}.
In the context of this work, it would be exciting to envision new strategies for sampling in a way which optimizes learning new information by ``sacrificing'' source-target similarity, but in such a way that we would know in advance that high quality DA is guaranteed.

\paragraph{Potential applications in COVID-19}
Researchers are diligently working to estimate both COVID-19 prevalence in given populations and the probability that individuals will be diagnosed with COVID-19 based on reported symptoms \cite{rao2020identification,imran2020ai4covid}.
Since the symptom distribution of this disease overlaps with a variety of other respiratory tract conditions, the above mentioned estimations are highly sensitive to the ratio of COVID-19 prevalence in the source and target datasets.
However, prevalence in the source is arbitrary and prevalence in the target may change by several orders of magnitude in short amounts of time.\\
An additional complication may arise from the fact that the conditional (reported) symptom distribution, given each condition, may change between source and target. Two plausible reasons are: 1) the source is based on physician reports while the target is based on user self-reports, and 2) as additional symptoms are associated with COVID-19, the populations' awareness of them increases over time (e.g., loss of the sense of smell and taste).\\
The scheme we have proposed in this paper can be used to generate reliable estimations, both on the population level and the individual level, under the given circumstances.

\section*{Acknowledgements}
The authors would like to thank Omer Sidis, Yael Steuerman, Amit Wolecki, Uri Shalit and Gabi Stanovsky for helpful discussions.

\bibliography{ref}

\begin{thebibliography}{10}

\bibitem{gonzalez2017review}
P.~Gonz{\'a}lez, A.~Casta{\~n}o, N.~V. Chawla, and J.~J.~D. Coz, ``A review on
  quantification learning,'' {\em ACM Computing Surveys (CSUR)}, vol.~50,
  no.~5, pp.~1--40, 2017.

\bibitem{mclachlan1988mixture}
G.~J. McLachlan and K.~E. Basford, {\em Mixture models: Inference and
  applications to clustering}, vol.~38.
\newblock M. Dekker New York, 1988.

\bibitem{mclachlan1992discriminant}
G.~McLachlan, ``Discriminant analysis and statistical pattern recognition,
  wiley,'' {\em New York}, 1992.

\bibitem{lipton2018detecting}
Z.~C. Lipton, Y.-X. Wang, and A.~Smola, ``Detecting and correcting for label
  shift with black box predictors,'' {\em arXiv preprint arXiv:1802.03916},
  2018.

\bibitem{redko2019advances}
I.~Redko, E.~Morvant, A.~Habrard, M.~Sebban, and Y.~Bennani, {\em Advances in
  Domain Adaptation Theory}.
\newblock Elsevier, 2019.

\bibitem{Kouw2019generalization}
W.~M. {Kouw} and M.~{Loog}, ``A review of domain adaptation without target
  labels,'' {\em IEEE Transactions on Pattern Analysis and Machine
  Intelligence}, pp.~1--1, 2019.

\bibitem{chidlovskii2019using}
B.~Chidlovskii, ``Using latent codes for class imbalance problem in
  unsupervised domain adaptation,'' {\em arXiv preprint arXiv:1909.08962},
  2019.

\bibitem{wu2010prediction}
J.~Wu, J.~Roy, and W.~F. Stewart, ``Prediction modeling using ehr data:
  challenges, strategies, and a comparison of machine learning approaches,''
  {\em Medical care}, pp.~S106--S113, 2010.

\bibitem{shickel2017deep}
B.~Shickel, P.~J. Tighe, A.~Bihorac, and P.~Rashidi, ``Deep ehr: a survey of
  recent advances in deep learning techniques for electronic health record
  (ehr) analysis,'' {\em IEEE journal of biomedical and health informatics},
  vol.~22, no.~5, pp.~1589--1604, 2017.

\bibitem{garg2020unified}
S.~Garg, Y.~Wu, S.~Balakrishnan, and Z.~C. Lipton, ``A unified view of label
  shift estimation,'' {\em arXiv preprint arXiv:2003.07554}, 2020.

\bibitem{forman2008quantifying}
G.~Forman, ``Quantifying counts and costs via classification,'' {\em Data
  Mining and Knowledge Discovery}, vol.~17, no.~2, pp.~164--206, 2008.

\bibitem{saerens2002adjusting}
M.~Saerens, P.~Latinne, and C.~Decaestecker, ``Adjusting the outputs of a
  classifier to new a priori probabilities: a simple procedure,'' {\em Neural
  computation}, vol.~14, no.~1, pp.~21--41, 2002.

\bibitem{csurka2017comprehensive}
G.~Csurka, ``A comprehensive survey on domain adaptation for visual
  applications,'' in {\em Domain adaptation in computer vision applications},
  pp.~1--35, Springer, 2017.

\bibitem{rao2020identification}
A.~S.~S. Rao and J.~A. Vazquez, ``Identification of covid-19 can be quicker
  through artificial intelligence framework using a mobile phone-based survey
  in the populations when cities/towns are under quarantine,'' {\em Infection
  Control \& Hospital Epidemiology}, pp.~1--18, 2020.

\bibitem{imran2020ai4covid}
A.~Imran, I.~Posokhova, H.~N. Qureshi, U.~Masood, S.~Riaz, K.~Ali, C.~N. John,
  and M.~Nabeel, ``Ai4covid-19: Ai enabled preliminary diagnosis for covid-19
  from cough samples via an app,'' {\em arXiv preprint arXiv:2004.01275}, 2020.

\end{thebibliography}
\bibliographystyle{ieeetr}

\end{document}